%% file: main.tex
\documentclass[%
 reprint,
 superscriptaddress,
 showkeys,
 longbibliography,
 aps,
 nofootinbib,
 amsmath,
 amssymb
]{revtex4-1}

\usepackage{lmodern}
\usepackage{enumitem}
\usepackage{graphicx}
\usepackage{dcolumn}
\usepackage{bm}
\usepackage{microtype}
\usepackage{tabularx}
\usepackage{booktabs}
\usepackage{multirow}
\usepackage{amsmath}
\usepackage{amssymb}
\usepackage{url}
\usepackage[table,xcdraw]{xcolor}
\usepackage[breaklinks, hidelinks, colorlinks=true, linkcolor=blue, citecolor=blue, urlcolor=blue, hypertexnames=false]{hyperref}

\newcommand{\medsam}{\textsc{MedSAM}}
\newcommand{\litemedsam}{\textsc{LiteMedSAM}}
\newcommand{\convnext}{\textsc{ConvNeXt}}
\newcommand{\convunext}{\textsc{ConvUNeXt}}
\newcommand{\lwnet}{\textsc{LW-Net}}
\newcommand{\thinrecall}{R_{\mathrm{thin}}}

\begin{document}

\title{Evaluating SAM-based and Convolutional Architectures in Thin Retinal Vessel Segmentation: Model Capacity, Computational Cost and Foundation Pre-training}

\author{Carlos Eduardo Fontaneli}
\email[]{carlos.fontaneli@estudante.ufscar.br}
\affiliation{Department of Computer Science, Federal University of S\~ao Carlos, Rod. Washington Lu\'is, km 235, 13565-905, S\~ao Carlos, SP, Brazil}

\author{Cesar H. Comin}
\email[Corresponding author: ]{comin@ufscar.br}
\affiliation{Department of Computer Science, Federal University of S\~ao Carlos, Rod. Washington Lu\'is, km 235, 13565-905, S\~ao Carlos, SP, Brazil}

\begin{abstract}
In retinal vessel segmentation, plain Vision Transformers (ViTs) are commonly expected to perform worse than convolutional neural networks (CNNs) on thin vessels, for reasons such as their coarse patch tokenization, their weak locality bias and their need for large training sets. Although this expectation is widely shared in the literature, how plain ViTs actually compare with CNNs on thin vessels has not been systematically quantified. We address this question by comparing a model with a plain ViT encoder (\medsam{}), a hybrid Transformer (\litemedsam{}) and three CNNs of different sizes on three retinal datasets (DRIVE, HRF, FIVES) and, as a second-domain check, a microscopy dataset (VessMap). We measured the centerline recall and precision of thin vessels and compared the models while also considering their computational cost and how many epochs they needed to train. We found that the pre-trained \medsam{} had a centerline recall of thin vessels that was not significantly different from that of the best CNN on all four datasets ($-0.6$ to $+2.5$ percentage points). A compact plain ViT obtained through architecture search but trained from scratch did not retain this performance. For pre-trained plain ViTs, our results do not support the expectation that they miss thin vessels, although similar thin-vessel performance can be obtained with much smaller models.

\end{abstract}


\maketitle

\input{sections/01_introduction}
\input{sections/02_related_work}
\input{sections/03_methodology}

\input{sections/04_results_discussion}

\input{sections/05_conclusion}
\input{sections/06_appendix}

\bibliography{references}

\end{document}

%% file: sections/01_introduction.tex
\section{Introduction}
\label{sec:introduction}

Segmentation of the retinal vessels in fundus images provides quantitative markers for the diagnosis and follow-up of diseases such as diabetic retinopathy and glaucoma~\cite{jin2022fives, moccia2018blood}. Thin vessels are a particular challenge in this task: they are only a few pixels wide, have low contrast, and contribute little to overlap metrics such as Dice, in which a wide vessel counts more pixels than a thin vessel of the same length.

Convolutional neural networks (CNNs) have dominated vessel segmentation since the U-Net~\cite{ronneberger2015u, siddique2021u, isensee2024nnunet, fadugba2024benchmarking, wittmann2025vesselfm}, and a carefully trained minimal U-Net, with far fewer parameters than the best published models, comes close to their results on retinal benchmarks~\cite{galdran2020littlewnetcouldstateoftheart}. Vision Transformers (ViTs)~\cite{dosovitskiy2020image} and the foundation models built on them, such as the Segment Anything Model~\cite{Kirillov_2023_ICCV} and its medical version \medsam{}~\cite{ma2024segment}, offer an alternative: large pre-trained encoders. These models are designed to be general-purpose and can be fine-tuned for specific tasks. A plain ViT, however, splits the image into fixed-size patches, $16 \times 16$ pixels in \medsam{}~\cite{ma2024segment}, and encodes each patch as one token. Its output is therefore a single low-resolution feature map, considered unsuitable for dense prediction~\cite{liu2021swin, wang2021pyramid, chen2023vitadapter}, and Transformers that encode such patches without convolutional features have been reported to lose the low-level detail needed for precise localization~\cite{chen2021transunet}. SAM-based models have been reported to struggle with thin and branching structures such as vessels~\cite{shi2023generalist, ke2023segment, ma2024segment, jiang2025gate}. This is a strong reason to expect a plain ViT to miss thin vessels. In addition, plain ViTs lack the locality bias of convolutions~\cite{dosovitskiy2020image, dascoli2021convit}: without large-scale pre-training they do not learn to attend locally in their early layers~\cite{raghu2021vision} and are more data-hungry than CNNs~\cite{liu2021efficient, valanarasu2021medical}. Vessel datasets, in contrast, have only tens to hundreds of annotated images.

To our knowledge, the thin-vessel performance of plain ViTs has not been systematically quantified or compared with that of CNNs. We therefore address three research questions:
\begin{enumerate}
    \item Can a plain ViT (\medsam{}) match the thin-vessel segmentation performance of CNNs, regardless of computational cost?
    \item How does a plain ViT compare with CNNs in thin-vessel performance relative to their computational cost?
    \item How much segmentation performance on thin vessels does foundation pre-training add to transformer-based models?
\end{enumerate}

To answer these questions, we trained the plain ViT \medsam{}, the hybrid Transformer \litemedsam{}, which combines convolutions and attention, and three CNNs of different sizes, \convnext{}~\cite{liu2022convnet}, \convunext{}~\cite{han2022convunext} and \lwnet{}~\cite{galdran2020littlewnetcouldstateoftheart}. We used three retinal datasets, DRIVE~\cite{Staal2004}, HRF~\cite{budai2013robust} and FIVES~\cite{jin2022fives}, and VessMap~\cite{viana2025new}, a microscopy dataset that we used to test the models in a second imaging domain. The two Transformers were also trained from scratch to quantify the contribution of foundation pre-training. In addition, we optimized the architecture of compact versions of the two Transformers on each dataset to test whether smaller models could reach the same performance.

Among our main results, fine-tuned \medsam{} had a centerline recall of thin vessels that was not significantly different from that of the best CNN on each dataset. Pre-training was essential for this performance, and a compact plain ViT found by architecture search and trained from scratch did not reach the performance of the pre-trained model. On HRF and FIVES, fine-tuned \medsam{} also had a significantly higher overall Dice than the best CNN.

%% file: sections/02_related_work.tex
\section{Related Work}
\label{sec:related_work}

Many deep segmentation networks build on fully convolutional networks~\cite{Long_2015_CVPR} and on the U-Net~\cite{ronneberger2015u}, whose skip connections carry high-resolution features to the decoder. U-Net variants have since become the primary tool for medical image segmentation~\cite{siddique2021u,isensee2024nnunet, fadugba2024benchmarking}. \convnext{}~\cite{liu2022convnet} modernized the CNN with large depthwise kernels and inverted bottlenecks. At the other end of the size range, Galdran et al.~\cite{galdran2020littlewnetcouldstateoftheart} showed that Little W-Net, a minimal cascaded U-Net with about 68,000 parameters, performs on par with or better than much larger models on retinal benchmarks when it is carefully trained and evaluated.

The ViT~\cite{dosovitskiy2020image} encodes an image as a sequence of patch tokens processed by self-attention~\cite{vaswani2017attention}. In medical segmentation, Transformers were first combined with CNNs, as in TransUNet~\cite{chen2021transunet}, or used in hierarchical forms, as in Swin-Unet~\cite{cao2022swin}. A systematic review of 36 studies that compared ViTs with CNNs in medical image analysis found that transformer-based models often performed better and that pre-training was important for them~\cite{takahashi2024comparison}, although the review was not specific to vessels. The Segment Anything Model (SAM)~\cite{Kirillov_2023_ICCV} is a promptable foundation model for segmentation, but its direct use on medical images is limited~\cite{zhang2024segmentmodelmedicalimage}. \medsam{}~\cite{ma2024segment} adapted SAM's ViT-B encoder to medical images by training it on over one million image--mask pairs. Its authors report that it has difficulty with vessel-like branching structures, which they attribute to the ambiguity of the box prompt~\cite{ma2024segment}. \litemedsam{}, the baseline of the CVPR~2024 challenge ``Segment Anything in Medical Images on Laptop'', distilled the encoder of \medsam{} into a TinyViT~\cite{wu2022tinyvit}, a lightweight hybrid of convolutional and Transformer layers~\cite{ma2024efficient}.

Several studies have examined how Transformers handle thin and small structures. In retinal vessel segmentation, hybrid models such as PCAT-UNet~\cite{chen2022pcatunet} add convolutions to Transformer blocks, arguing that self-attention captures global context but not the local detail needed for microvessels. A comparison of eight convolutional and Transformer-based U-shaped networks on four retinal datasets found small overall differences, with the CNNs ahead on the low-resolution DRIVE images and the Transformers ahead on the high-resolution HRF images~\cite{sadrabadi2024comparative}. For small objects in general, the segmentation performance of ViTs depends on the patch size, as shown for tumors of different sizes~\cite{mojtahedi2022towards}. Segmentation foundation models show similar difficulties. Zero-shot SAM fails on retinal vessels~\cite{shi2023generalist} and often neglects thin object structures in natural images~\cite{ke2023segment}. Zhang et al.~\cite{zhang2026quantifying} found that SAM, SAM~2 and HQ-SAM perform worse on objects with dense, tree-like morphology, such as retinal vessels. They attributed this to patches that contain mostly boundary pixels and few interior pixels of the object, an imbalance that grows with the model's effective patch size, and found that fine-tuning does not remove this limitation. FineSAM++~\cite{jiang2025gate} addresses such failures on vessels by refining the output of SAM in uncertain regions.

Thin vessels have also been treated as a separate target in training and evaluation. Because pixel-wise losses are dominated by the thick vessels, Yan et al.~\cite{yan2018joint} proposed a segment-level loss that gives thin vessels more weight. The clDice loss~\cite{Shit_2021_CVPR} for tubular structures is built on the skeleton-based sensitivity and precision of a segmentation, which correspond to the centerline recall and precision used in this work. A recent study that stratified sensitivity by vessel width found that the detection of thin vessels can vary considerably with the input resolution while Dice remains stable~\cite{amarnath2026effect}. These studies evaluated losses, metrics or input resolution, mostly with CNNs. To our knowledge, no study has compared foundation models against CNNs with thin vessels evaluated separately from the rest of the vessel tree. Moreover, comparisons between architectures often do not optimize the hyperparameters of every model with the same effort, which can favor the proposed model over poorly configured baselines~\cite{isensee2024nnunet}.

%% file: sections/03_methodology.tex
\section{Methodology}
\label{sec:methodology}

\subsection{Datasets}
\label{subsec:method_problem_data}

We used three retinal fundus datasets, DRIVE~\cite{Staal2004}, HRF~\cite{budai2013robust} and FIVES~\cite{jin2022fives}, and one microscopy dataset, VessMap~\cite{viana2025new}. They were chosen to differ in two properties related to the expected weaknesses of a plain ViT (Section~\ref{sec:introduction}): the number of training images, from 16 to 600, and the image size, from 256 to 3504 pixels per side, which sets how many pixels a vessel spans relative to a patch.

DRIVE is the standard benchmark of retinal vessel segmentation~\cite{moccia2018blood}, which allows our results to be related to the literature. It has 40 images of $565 \times 584$ pixels with an official split of 20 training and 20 test images. We held out four of the training images (images 37 to 40) for validation and trained on the other 16. DRIVE is therefore the dataset with the fewest training images, the setting in which a plain ViT trained from scratch is expected to suffer most. HRF has 45 images of $3504 \times 2336$ pixels, 15 each of healthy eyes and of eyes with diabetic retinopathy or glaucoma~\cite{budai2013robust}. FIVES has 800 images of $2048 \times 2048$ pixels, 200 each of normal eyes and of eyes with age-related macular degeneration, diabetic retinopathy or glaucoma, with an official split of 600 training and 200 test images. With 600 training images, FIVES tests whether the gain from foundation pre-training becomes smaller when many training images are available.

VessMap has 100 confocal fluorescence microscopy images ($256 \times 256$ pixels) of the vasculature of the mouse brain cortex, with low contrast. They were selected from a larger set of images to be as heterogeneous as possible in contrast, noise, vessel density and the variation of intensity along the vessels~\cite{viana2025new}. VessMap tests whether our findings hold in a second imaging domain.

HRF and VessMap have no official split, so we split each at random, with a fixed seed, into 70\% training and 30\% test images. On HRF, FIVES and VessMap, no separate validation set was held out, and checkpoint selection, early stopping and the hyperparameter search used the test images. This protocol was the same for every model.

\subsection{Models}
\label{subsec:method_models}
\label{subsec:method_matrix}

Table~\ref{tab:model_matrix} summarizes the architectures used. \medsam{}~\cite{ma2024segment} is the plain ViT, a single-scale ViT without convolutional stages. It uses the ViT-B encoder of SAM, with $16 \times 16$~px patches and a $1024 \times 1024$~px input, and was trained on over one million medical image--mask pairs. Its training data include fundus images, but only from optic disc and cup datasets, and none of our four datasets is among them~\cite{ma2024segment}. \litemedsam{}~\cite{ma2024efficient} is the hybrid model. Its TinyViT encoder~\cite{wu2022tinyvit} combines convolutional stages (mobile inverted bottleneck convolutions) with windowed attention and has a $256 \times 256$~px input. It was distilled and fine-tuned on the same data as \medsam{}~\cite{ma2024efficient}. Both transformer-based models keep SAM's prompt encoder and mask decoder~\cite{Kirillov_2023_ICCV}. In training and inference, we set the prompt as a fixed box covering the whole input window, and the prompt encoder was not trained. The decoder upsamples the $64 \times 64$ grid of encoder tokens by a factor of four. \medsam{} therefore predicts at a quarter of its input resolution, and its logits are upsampled bilinearly to the size of the window. \litemedsam{}, whose encoder reduces its 256~px window to the same grid, predicts at its input resolution. The three CNNs span more than three orders of magnitude in size: \convnext{}, a ConvNeXt-Large encoder~\cite{liu2022convnet} with a transposed-convolution decoder; \convunext{}, a U-Net built from ConvNeXt-style blocks with attention gates~\cite{han2022convunext}; and \lwnet{}, our implementation of Little W-Net~\cite{galdran2020littlewnetcouldstateoftheart}.

\begin{table}[tb]
\caption{The five architectures and their parameters. \convunext{} and \lwnet{} are U-Nets without a separate, named encoder, so for them the encoder column gives the type of block of the encoding path. The size of \convunext{} depends on the dataset (Appendix~\ref{app:training}).}
\label{tab:model_matrix}
\begin{ruledtabular}
\begin{tabular}{lllr}
Model & Type & Encoder & Params (M) \\
\midrule
\medsam{} & Plain ViT & ViT-B & 93.7 \\
\litemedsam{} & Hybrid & TinyViT & 9.8 \\
\convnext{} & CNN & ConvNeXt-L & 211.4 \\
\convunext{} & CNN & ConvNeXt blocks & 3.5 / 13.8 \\
\lwnet{} & CNN & Conv.\ blocks & 0.07 \\
\end{tabular}
\end{ruledtabular}
\end{table}

\subsection{Training protocol}
\label{subsec:method_training}

We trained the three CNNs from scratch. The two transformer-based models were trained in two ways: fine-tuned from their published pre-trained weights~\cite{ma2024segment, ma2024efficient}, with all weights updated except those of the prompt encoder, and trained from scratch. Comparing pre-trained transformer-based models with CNNs trained from scratch was deliberate, because these are the configurations in which each kind of model is usually applied. Compact segmentation CNNs are trained from scratch~\cite{ronneberger2015u, galdran2020littlewnetcouldstateoftheart, han2022convunext, liu2022frunet, isensee2021nnunet}, while transformer-based foundation models are fine-tuned from their pre-trained weights. \convnext{} was included as a CNN in the same cost class as \medsam{}. Although ImageNet weights exist for its encoder, pre-trained weights are rarely available for compact CNNs, so \convnext{} followed the same protocol as the other two CNNs. Training the transformer-based models also from scratch separates the effect of pre-training from that of the architecture.

For each configuration and dataset, a search of 100 trials chose the learning rate, the weight decay, the batch size, the base width of \convunext{} and \lwnet{} and the stochastic-depth rate of \convnext{}. Table~\ref{tab:search_space} contains the full search space. We ran the searches with Optuna~\cite{akiba2019optuna} (version 4.7.0), using its tree-structured Parzen estimator sampler~\cite{bergstra2011algorithms} and median pruning on the validation Dice. In total, the 36 searches ran 3,602 trials in 2,093 GPU hours. The CNNs were trained on an NVIDIA RTX 3090 GPU (24~GB) and the transformer-based models on NVIDIA RTX PRO 5000 (48~GB) and RTX PRO 4000 (24~GB) GPUs.

All models were trained with the soft Dice loss~\cite{milletari2016vnet} and AdamW~\cite{loshchilov2019decoupled}, in PyTorch~\cite{paszke2019pytorch} (version 2.10.0) with PyTorch Lightning~\cite{falcon2026lightning} (version 2.6.1).  We kept the trial with the highest validation Dice. Training stopped after 50 epochs without improvement of the validation Dice.

Each training sample was a random crop of the size of the model's input (256~px, or 1024~px for \medsam{}), taken after random flips and rotations. Images smaller than the input, which is the case of the DRIVE and VessMap images for \medsam{}, were upscaled to it, so \medsam{} saw DRIVE at about 1.75 times and VessMap at 4 times their native scale. Appendix~\ref{app:training} gives the full training details.

As an additional investigation, we tested whether smaller transformer-based models could reach the performance of the original ones. We trained, from scratch, a compact version of each transformer-based model, with a 256~px input and an architecture optimized on each dataset. For these models, the search of 100 trials covered the architecture in addition to the training hyperparameters. For compact \medsam{}, the patch size, the embedding dimension, the depth, the number of attention heads, the MLP ratio and the attention window were optimized. For compact \litemedsam{}, the embedding dimensions, the depths of its stages, the number of heads, the attention windows and the MLP ratio were optimized. Appendix~\ref{app:training} describes the search space, and Table~\ref{tab:compact_models} contains the selected architectures.

\subsection{Evaluation protocol and metrics}
\label{subsec:method_metrics}

We evaluated every model on the full test images at native resolution with the sliding-window inference of MONAI~\cite{cardoso2022monai} (version 1.5.2). The windows had the size of the model's input with an overlap of 25\% and they were combined with Gaussian weighting. Images smaller than a window were padded to it. The sigmoid output was binarized at 0.5. Because \medsam{} was trained on upscaled DRIVE and VessMap images, on these two datasets we upscaled each test image to its $1024 \times 1024$~px input, resized the logits back to the native resolution with bilinear interpolation, and scored them against the native ground truth. We call this the upscaled protocol, and the alternative, in which the native image is padded to the window, the padded protocol. The upscaled protocol evaluates \medsam{} at its training scale, and we used it for every \medsam{} result except the comparison of the two protocols.

All metrics were computed for each test image from the binarized prediction $\hat{Y}$ and the ground truth $Y$. We used the Dice score over the full image as an overall measure of performance. The centerline metrics use the skeleton $S(\cdot)$ of a mask, computed with the \texttt{skeletonize} function of scikit-image~\cite{vanderwalt2014scikit} (version 0.26.0), which implements Zhang--Suen thinning~\cite{zhang1984fast}. Our main metrics are based on the centerline recall,
\begin{equation}
R_{\mathrm{cl}} = \frac{|S(Y) \cap \hat{Y}|}{|S(Y)|},
\end{equation}
the fraction of ground-truth centerline pixels that fall inside the predicted mask. To check that high values of $R_{\mathrm{cl}}$ were not obtained at the cost of more false positives, we also computed the centerline precision,
\begin{equation}
P_{\mathrm{cl}} = \frac{|S(\hat{Y}) \cap Y|}{|S(\hat{Y})|},
\end{equation}
the fraction of predicted centerline pixels that fall inside the ground-truth mask. $R_{\mathrm{cl}}$ and $P_{\mathrm{cl}}$ were used in \cite{Shit_2021_CVPR} to define the clDice metric.

To measure recall as a function of vessel caliber, we grouped the ground-truth centerline pixels by radius into the bins $[0.5, 1.5)$, $[1.5, 2.5)$, $[2.5, 3.5)$, $[3.5, 4.5)$ and $[4.5, \infty)$~px. The radius of a centerline pixel was taken as the value of the Euclidean distance transform of the mask at that pixel. We call a vessel thin when its radius is below 1.5~px. The centerline recall of thin vessels is the recall of the first bin,
\begin{equation}
    \thinrecall = \frac{|S_{\mathrm{thin}}(Y) \cap \hat{Y}|}{|S_{\mathrm{thin}}(Y)|},
    \label{eq:thin_recall}
\end{equation}
where $S_{\mathrm{thin}}(Y)$ is the set of ground-truth centerline pixels with radius below 1.5~px. The centerline precision of thin vessels,
\begin{equation}
    P_{\mathrm{thin}} = \frac{|S_{\mathrm{thin}}(\hat{Y}) \cap Y|}{|S_{\mathrm{thin}}(\hat{Y})|},
\end{equation}
considers only the predicted centerline pixels whose radius, measured in the prediction, is below 1.5~px. The two thin-vessel metrics are therefore computed on different sets of pixels: $\thinrecall$ on the ground-truth thin centerline and $P_{\mathrm{thin}}$ on the predicted one. When a denominator is empty, the metric is undefined for that image and is left out of the mean.

Note that $P_{\mathrm{thin}}$ is used in our analysis only as a sanity check to quantify if models are oversegmenting the vessels to improve their $\thinrecall$. This is because $P_{\mathrm{thin}}$ cannot be considered a cooutberbalance to $\thinrecall$ the same way that precision is to recall. $\thinrecall$ measures the capacity of a model to detect thin vessels present in the ground-truth, while $P_{\mathrm{thin}}$ measures false positives that happen to be thin structures. For instance, a model that predicts vessel for all pixels will have a $\thinrecall$ of 1, but undefined $P_{\mathrm{thin}}$, because it will have no predicted thin centerline pixels.

We counted the parameters and the floating-point operations of each trained checkpoint with fvcore~\cite{fvcore2022} (version 0.1.5), counting one multiply--add as one FLOP. As a measure of training difficulty, we report the number of epochs the selected run needed to first reach 95\% of its best validation Dice.

Significance tests used a two-sided Wilcoxon signed-rank test~\cite{wilcoxon1945individual} on per-image paired differences, with the Holm correction~\cite{holm1979simple}. A difference is called significant when its Holm-adjusted p-value is below 0.05. The result tables report the mean $\pm$ standard deviation of each metric over the test images. Bold marks the best value of a comparison, underlining marks the values not significantly different from the best one, and an asterisk marks a significant difference in a pairwise comparison.

%% file: sections/04_results_discussion.tex
\section{Results}
\label{sec:results_discussion}

\subsection{Overall segmentation performance}
\label{subsec:results_overall}

\input{figures/results/tab01_global_baseline}

\begin{figure*}[tbp]
\centering
\includegraphics[width=0.9\textwidth]{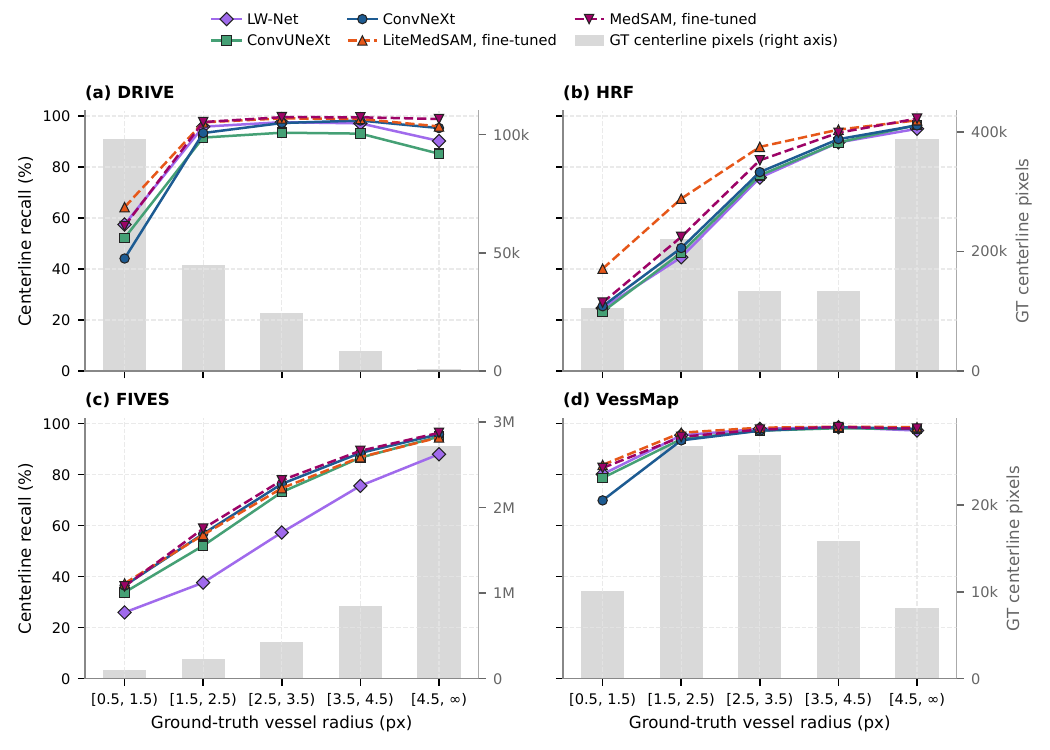}
\caption{Mean centerline recall for different vessel radius ranges. The first range, radius in $[0.5, 1.5)$~px, shows $\thinrecall$. Grey bars show the number of ground-truth centerline pixels for each range, calculated on the test images.}
\label{fig:recall_caliber}
\end{figure*}

Table~\ref{tab:global_baseline} reports the Dice, the centerline recall $R_{\mathrm{cl}}$ and the centerline precision $P_{\mathrm{cl}}$ of the five models, computed over all vessels. The p-values of the pairwise tests are given in Appendix~\ref{app:statistics}, in Table~\ref{tab:pvalues} for the Dice and in Table~\ref{tab:pvalues_other} for $R_{\mathrm{cl}}$ and $P_{\mathrm{cl}}$. The best model of each retinal dataset reached a Dice close to published values: 82.3\% on DRIVE (\litemedsam{}) and 81.7\% on HRF (\medsam{}), against 82.8\% and 81.0\% reported for Little W-Net~\cite{galdran2020littlewnetcouldstateoftheart}, and 92.2\% on FIVES (\medsam{}), against 90.4\% reported for FR-UNet~\cite{liu2022frunet} by Fadugba et al.~\cite{fadugba2024benchmarking}. VessMap has no published Dice to compare with. The \lwnet{} stayed 1.7 and 4.1 points below the published values on DRIVE and HRF~\cite{galdran2020littlewnetcouldstateoftheart}. This might be because the published protocol optimized the threshold value on the training set and evaluated the Dice score only inside the field of view.

\begin{figure*}[tbp]
\centering
\includegraphics[width=\textwidth]{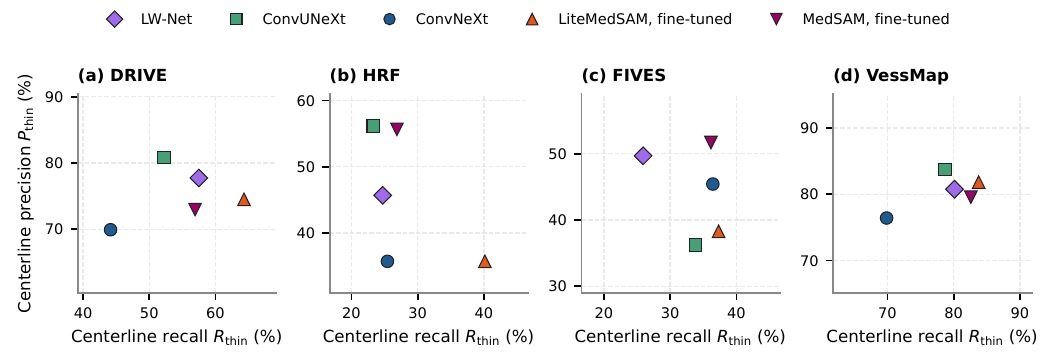}
\caption{Centerline precision against centerline recall of thin vessels, per dataset.}
\label{fig:thin_pr}
\end{figure*}

No model had the highest mean on every dataset and metric. For the Dice score, however, an attention-based model had the highest mean on every dataset: \litemedsam{} on DRIVE and VessMap, and \medsam{} on HRF and FIVES. On VessMap, only \convnext{} was significantly lower than the other models.

These overall metrics say little about thin vessels. $R_{\mathrm{cl}}$ and $P_{\mathrm{cl}}$ weigh every centerline pixel equally. Thin vessels hold 55.3\% of the ground-truth centerline pixels of DRIVE and 2.3 to 11.7\% of those of the other three datasets (pixel counts are shown in Fig.~\ref{fig:recall_caliber}). Furthermore, for the Dice score, a wide vessel weighs more than a thin vessel of the same length.

\subsection{Segmentation performance on thin vessels}
\label{subsec:results_thin}

Fig.~\ref{fig:recall_caliber} shows the mean centerline recall as a function of vessel radius. The compact models are not included because the figure focus on raw performance and also to reduce visual clutter. For every model and dataset, the recall was lowest in the first bin, radius below 1.5~px. On HRF and FIVES, the recall fell steadily as the radius decreased, while on DRIVE and VessMap it was nearly constant above 1.5~px and dropped only in the first bin. The recall of the first bin is the centerline recall of thin vessels, $\thinrecall$, compared below.

Fine-tuned \medsam{} had an $\thinrecall$ within the range of the three CNNs, that is, between their lowest and highest means, on DRIVE and FIVES, and above this range on HRF and VessMap. Its $\thinrecall$ was not significantly different from that of the best CNN, the CNN with the highest mean for the same dataset and metric, on any dataset. The hybrid \litemedsam{} had the highest mean $\thinrecall$ on every dataset, significantly above every CNN on DRIVE and HRF. Table~\ref{tab:pvalues} of Appendix~\ref{app:statistics} gives the p-values of these comparisons.

A model can raise its $\thinrecall$ by over-segmenting, that is, by trading precision for recall. To check whether this was the case for any of the models, Fig.~\ref{fig:thin_pr} plots the centerline precision of thin vessels, $P_{\mathrm{thin}}$, against $\thinrecall$. On DRIVE, \medsam{} had an $\thinrecall$ not significantly different from those of \lwnet{} and \convunext{}, but a significantly lower $P_{\mathrm{thin}}$. On HRF, \medsam{} and \convunext{} did not differ significantly in either metric. On FIVES, no model was significantly better than \medsam{} in either metric, and each of the other four models was significantly worse in at least one. On VessMap, \medsam{} and \lwnet{} did not differ significantly in either metric.

Taken together, the results indicate that the high values of $\thinrecall$ of \medsam{} did not come at the cost of a low $P_{\mathrm{thin}}$, when compared to CNNs with similar performance. The model had equivalent performance to a CNN on HRF and VessMap and a better overall performance on FIVES.

For \litemedsam{}, the higher $\thinrecall$ came with a lower $P_{\mathrm{thin}}$ than that of the best CNN on every dataset, with a significant difference on DRIVE, HRF and FIVES.

\begin{figure*}[tbp]
\centering
\includegraphics[width=\textwidth]{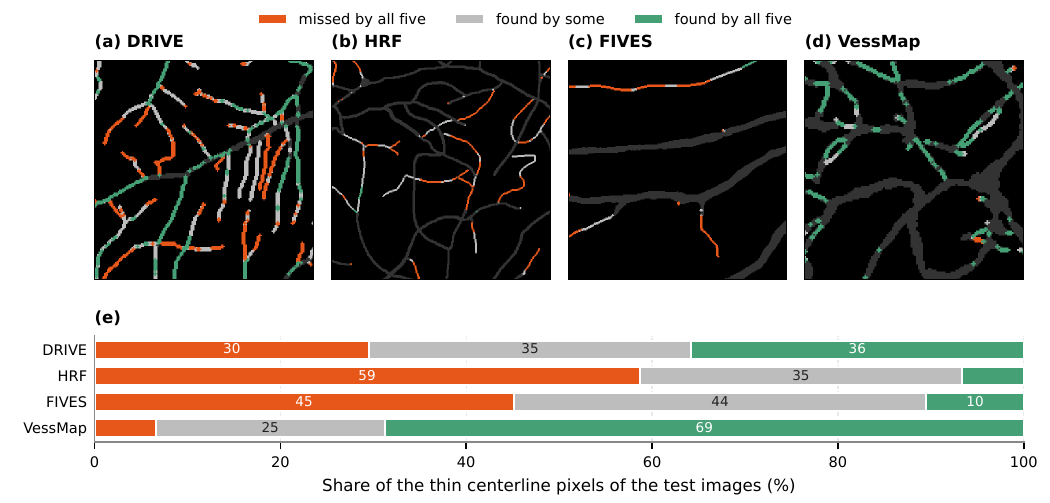}
\caption{Thin vessel detection agreement among the models. (a--d) Window of each test set with the most thin centerline pixels. Non-thin vessels are shown in dark grey. (e) Agreements between the models over all test images.}
\label{fig:thin_errors}
\end{figure*}

\begin{figure*}[tbp]
\centering
\includegraphics[width=\textwidth]{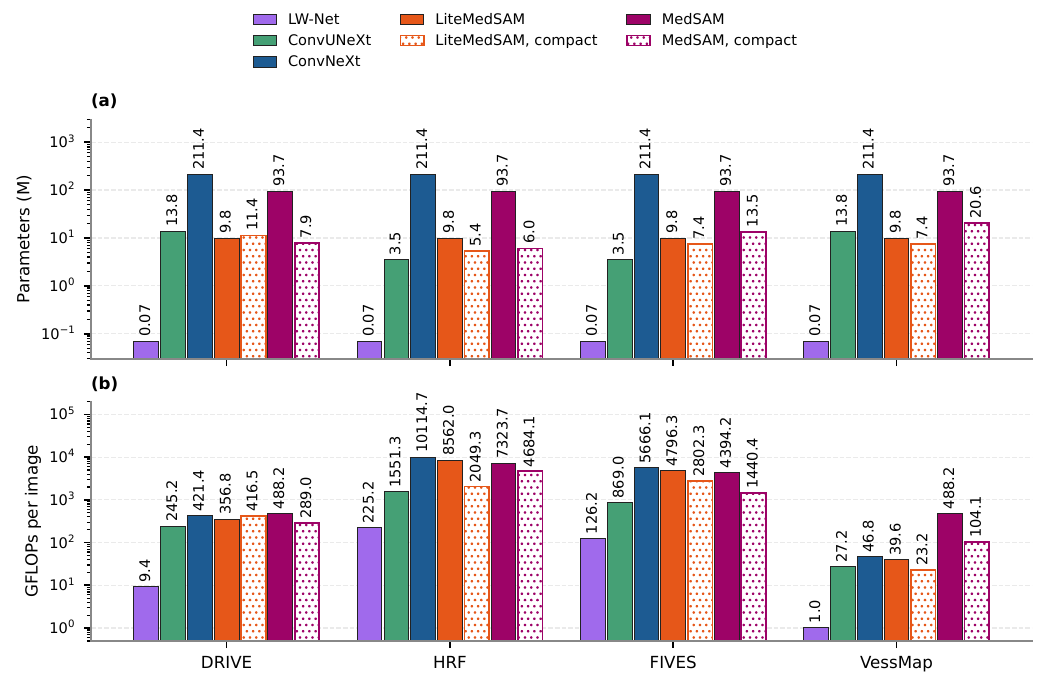}
\caption{Cost of each architecture. (a)~parameters and (b)~GFLOPs per test image, log scale. }
\label{fig:cost}
\end{figure*}

Fig.~\ref{fig:thin_errors} shows examples of thin vessels found and missed by the five models, and the share of thin centerline pixels missed by all five, found by some and found by all five. The share on which the models disagreed was 35\% on DRIVE and HRF, higher on FIVES (44\%) and lower on VessMap (25\%). The share missed by all five models varied much more, from 59\% on HRF and 45\% on FIVES to 30\% on DRIVE and less than 10\% on VessMap. In the examples, the models seem to miss or disagree more on terminal branches. A possible reason is that a thin segment that is not terminal have better support for segmentation when there are vessels connected to both of its ends.

\subsection{Model cost and compact architectures}
\label{subsec:results_cost}

\begin{figure*}[tbp]
\centering
\includegraphics[width=\textwidth]{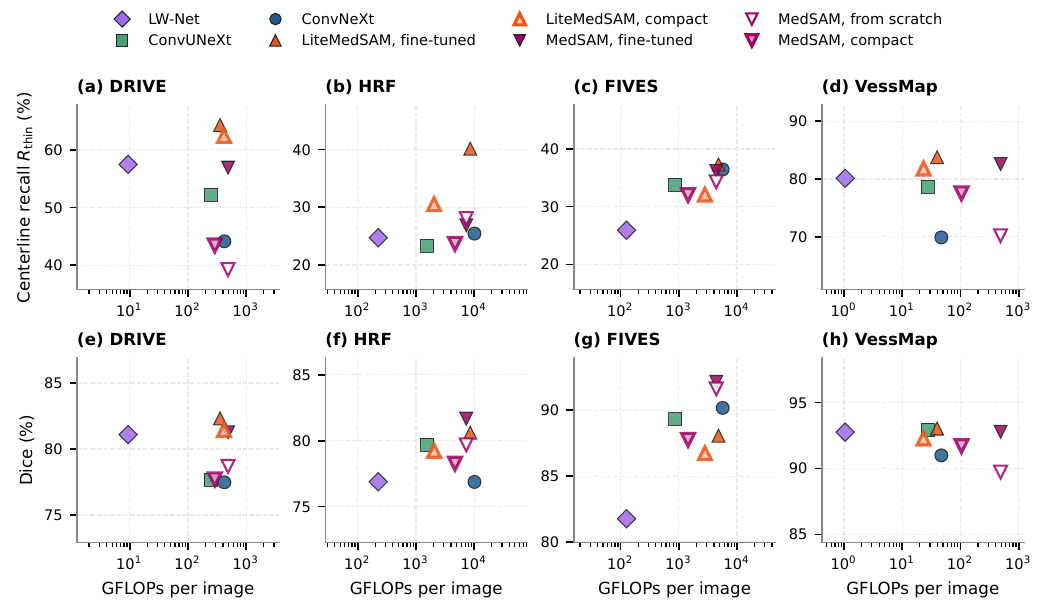}
\caption{Relationship between performance and cost. $\thinrecall$ (a--d) and Dice (e--h) against GFLOPs per test image.}
\label{fig:compact_tradeoff}
\end{figure*}

\begin{figure*}[tbp]
\centering
\includegraphics[width=\textwidth]{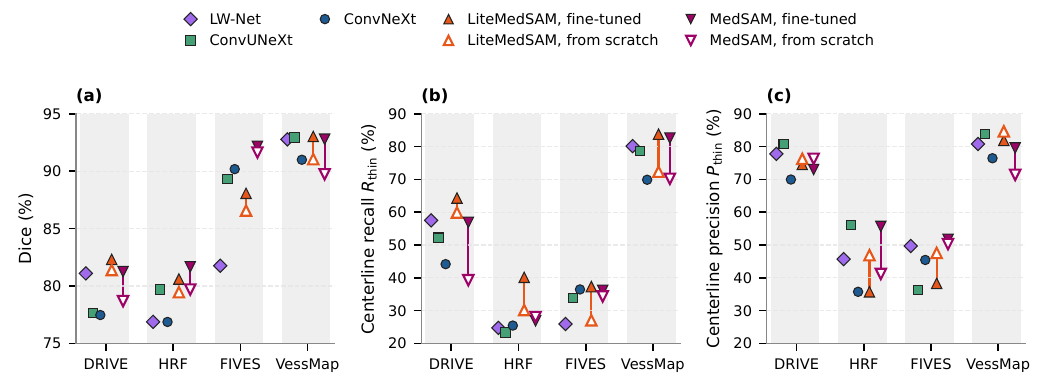}
\caption{Effect of pre-training. (a)~Dice, (b)~$\thinrecall$ and (c)~$P_{\mathrm{thin}}$ of the two transformer-based models fine-tuned (filled) and trained from scratch (hollow), next to the CNNs.}
\label{fig:pretraining}
\end{figure*}

Fig.~\ref{fig:cost} shows the parameters and the GFLOPs per test image of each architecture. \medsam{} had the second-largest parameter count, after \convnext{}, and the highest cost per image on DRIVE and VessMap, whose images are smaller than its 1024~px window and are upscaled to it. On the larger HRF and FIVES images, the models with 256~px windows need many more windows per image than \medsam{}, so \convnext{} and \litemedsam{} cost more per image than \medsam{}.

By parameter count, the models fall into three cost classes: \medsam{} and \convnext{}; \litemedsam{}, \convunext{} and the compact models; and \lwnet{}. Models of the same class can be compared fairly, while a comparison across classes is fair only when the aim is the highest accuracy regardless of cost. The search for compact \litemedsam{}, which was not limited to models smaller than \litemedsam{}, returned models of the same order of size and cost as \litemedsam{}, and on DRIVE a slightly larger model (Table~\ref{tab:compact_models}).

\begin{figure*}[tbp]
\centering
\includegraphics[width=\textwidth]{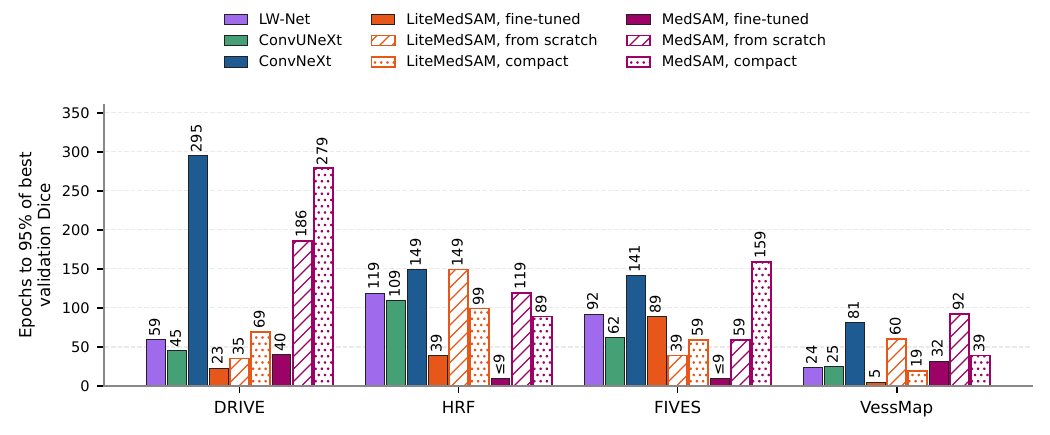}
\caption{Training epochs required to reach 95\% of the best validation Dice. Models were validated every epoch or every 10 epochs (see Table~\ref{tab:search_space} for details). ``$\le$'' means that the model reached the performance at the first validation. }
\label{fig:convergence}
\end{figure*}

\input{figures/results/tab02_padding_protocol}

Fig.~\ref{fig:compact_tradeoff} plots $\thinrecall$ and Dice against the GFLOPs per test image. Fine-tuned \medsam{}, whose $\thinrecall$ was not significantly different from that of the best CNN on any dataset, belongs to the most expensive class by parameter count. Its compact version did not reach the same performance. Against the best CNN, its $\thinrecall$ was significantly lower on DRIVE and FIVES, and its $P_{\mathrm{thin}}$ and Dice were significantly lower on all four datasets.

The search for compact \medsam{} selected finer patches than the 16~px of \medsam{}. On DRIVE, HRF and VessMap, the optimal patch size was 4~px, while on FIVES it was 8~px. Full details are shown in Table~\ref{tab:patch_size}. Patches of 1 and 2~px ran out of GPU memory, so 4~px was the finest patch that could be trained.

\subsection{Pre-training and convergence speed}
\label{subsec:results_pretraining}

Fig.~\ref{fig:pretraining} compares \medsam{} and \litemedsam{} fine-tuned from their pre-trained weights with the same architectures trained from scratch, next to the three CNNs. Table~\ref{tab:pretraining_gain} in Appendix~\ref{app:results_detail} provides the values used in the figure for \medsam{} and \litemedsam{}.

Pre-training raised the Dice of \medsam{} significantly on every dataset. Its effect on the $\thinrecall$ of \medsam{} depended on the dataset. $\thinrecall$ rose by 17.6 and 12.4 points on DRIVE and VessMap, by only 1.9 points on FIVES, and did not change significantly on HRF. Without pre-training, the $\thinrecall$ of \medsam{} was significantly lower than that of the best CNN on DRIVE, FIVES and VessMap. Pre-training also raised the $P_{\mathrm{thin}}$ of \medsam{} on HRF and VessMap, and slightly reduced it on DRIVE. For the hybrid \litemedsam{}, pre-training raised $\thinrecall$ significantly on every dataset.

For \medsam{}, the Dice gain from pre-training was smallest on FIVES, the dataset with the most training images. Interestingly, the gain in $\thinrecall$ for \medsam{} was lower on the high-resolution datasets.

Fig.~\ref{fig:convergence} shows the number of epochs until each selected run first reached 95\% of its best validation Dice. Pre-trained \medsam{} converged the fastest of all models on HRF and FIVES, where it reached this level at the first validation, and about as fast as \lwnet{} and \convunext{} on DRIVE and VessMap. Without pre-training, \medsam{} needed more epochs on every dataset, and so did compact \medsam{}. Pre-trained \litemedsam{} converged faster than every CNN, except on FIVES, where it was slower than \convunext{}.

\subsection{Inference protocol of MedSAM: padding versus upscaling}
\label{subsec:results_padding}

Upscaling the DRIVE and VessMap images to the 1024~px window of \medsam{} adds no information, but it changes the scale at which the model sees the thin vessels. To measure the effect of this choice, we compared the upscaled and padded protocols of Section~\ref{subsec:method_metrics} on these two datasets. The results are shown in Table~\ref{tab:padding}.

The padded protocol gave a significantly lower Dice and $\thinrecall$ than the upscaled protocol, both for fine-tuned \medsam{} and for \medsam{} trained from scratch. For fine-tuned \medsam{}, $\thinrecall$ fell by 17.9 points on DRIVE and 41.3 points on VessMap, below the range of the CNNs on both datasets.

%% file: figures/results/tab01_global_baseline.tex
\begin{table*}[tbp]
\caption{Overall segmentation performance. Values in bold indicate the best performance. Underlined values are not significantly different from the best (bold) of the same row.}
\label{tab:global_baseline}
\begin{ruledtabular}
\begin{tabular}{llccccc}
 & & \multicolumn{3}{c}{CNN} & \multicolumn{2}{c}{Fine-tuned} \\
\cmidrule(lr){3-5} \cmidrule(lr){6-7}
Dataset & Metric & \lwnet{} & \convunext{} & \convnext{} & \litemedsam{} & \medsam{} \\
\midrule
\multirow{3}{*}{\shortstack{DRIVE\\($n=20$)}} & Dice & $81.1 \pm 2.4$ & $77.6 \pm 15.1$ & $77.5 \pm 2.5$ & {\boldmath$82.3 \pm 1.9$} & $81.3 \pm 1.3$ \\
 & $R_{\mathrm{cl}}$ & $74.9 \pm 7.2$ & $70.1 \pm 16.7$ & $66.9 \pm 7.5$ & {\boldmath$79.4 \pm 7.1$} & $75.5 \pm 6.9$ \\
 & $P_{\mathrm{cl}}$ & \underline{$89.3 \pm 2.3$} & {\boldmath$90.5 \pm 4.4$} & \underline{$89.5 \pm 3.1$} & $87.6 \pm 2.5$ & $87.3 \pm 2.9$ \\
\midrule
\multirow{3}{*}{\shortstack{HRF\\($n=14$)}} & Dice & $76.9 \pm 6.9$ & $79.7 \pm 5.1$ & $76.9 \pm 6.2$ & \underline{$80.6 \pm 5.9$} & {\boldmath$81.7 \pm 4.5$} \\
 & $R_{\mathrm{cl}}$ & $72.0 \pm 12.0$ & $72.8 \pm 9.5$ & $73.8 \pm 7.6$ & {\boldmath$82.2 \pm 7.6$} & $76.6 \pm 8.6$ \\
 & $P_{\mathrm{cl}}$ & \underline{$84.0 \pm 11.5$} & \underline{$88.5 \pm 7.6$} & $79.5 \pm 10.0$ & $83.9 \pm 8.0$ & {\boldmath$88.9 \pm 6.7$} \\
\midrule
\multirow{3}{*}{\shortstack{FIVES\\($n=200$)}} & Dice & $81.8 \pm 14.7$ & $89.3 \pm 9.2$ & $90.2 \pm 8.5$ & $88.1 \pm 9.8$ & {\boldmath$92.2 \pm 8.5$} \\
 & $R_{\mathrm{cl}}$ & $78.7 \pm 17.3$ & $87.4 \pm 11.1$ & $88.9 \pm 10.0$ & $87.6 \pm 10.9$ & {\boldmath$89.6 \pm 10.4$} \\
 & $P_{\mathrm{cl}}$ & $88.0 \pm 11.1$ & $92.1 \pm 7.4$ & $92.1 \pm 7.7$ & $89.4 \pm 10.6$ & {\boldmath$95.4 \pm 4.7$} \\
\midrule
\multirow{3}{*}{\shortstack{VessMap\\($n=30$)}} & Dice & \underline{$92.8 \pm 2.4$} & \underline{$92.9 \pm 2.1$} & $91.0 \pm 2.3$ & {\boldmath$93.0 \pm 2.0$} & \underline{$92.8 \pm 2.1$} \\
 & $R_{\mathrm{cl}}$ & $95.9 \pm 3.6$ & $95.4 \pm 4.2$ & $94.7 \pm 4.1$ & {\boldmath$96.9 \pm 3.1$} & \underline{$96.5 \pm 3.7$} \\
 & $P_{\mathrm{cl}}$ & \underline{$96.2 \pm 3.1$} & {\boldmath$97.0 \pm 2.0$} & \underline{$96.3 \pm 3.1$} & \underline{$96.8 \pm 2.8$} & $96.3 \pm 2.7$ \\
\end{tabular}
\end{ruledtabular}
\end{table*}

%% file: figures/results/tab02_padding_protocol.tex
\begin{table*}[tbp]
\caption{\medsam{} evaluated with the padded and the upscaled protocol. Bold indicates the best value. Significant differences are indicated by an asterisk.}
\label{tab:padding}
\begin{ruledtabular}
\begin{tabular}{llcccccc}
 & & \multicolumn{3}{c}{DRIVE ($n=20$)} & \multicolumn{3}{c}{VessMap ($n=30$)} \\
\cmidrule(lr){3-5} \cmidrule(lr){6-8}
\medsam{} & Metric & Padded & Upscaled & $\Delta$ & Padded & Upscaled & $\Delta$ \\
\midrule
\multirow{3}{*}{Fine-tuned} & Dice & $75.2 \pm 1.8$ & {\boldmath$81.3 \pm 1.3$} & $+6.1^{*}$ & $78.7 \pm 7.5$ & {\boldmath$92.8 \pm 2.1$} & $+14.1^{*}$ \\
 & $\thinrecall$ & $39.0 \pm 10.9$ & {\boldmath$56.9 \pm 11.8$} & $+17.9^{*}$ & $41.3 \pm 12.7$ & {\boldmath$82.6 \pm 14.1$} & $+41.3^{*}$ \\
 & $P_{\mathrm{thin}}$ & $58.8 \pm 7.9$ & {\boldmath$73.0 \pm 6.1$} & $+14.2^{*}$ & $76.8 \pm 13.2$ & {\boldmath$79.6 \pm 14.8$} & $+2.8$ \\
\midrule
\multirow{3}{*}{From scratch} & Dice & $65.7 \pm 3.0$ & {\boldmath$78.7 \pm 1.6$} & $+13.0^{*}$ & $68.2 \pm 6.4$ & {\boldmath$89.7 \pm 3.7$} & $+21.5^{*}$ \\
 & $\thinrecall$ & $25.3 \pm 8.0$ & {\boldmath$39.3 \pm 11.2$} & $+14.0^{*}$ & $43.6 \pm 19.2$ & {\boldmath$70.2 \pm 20.9$} & $+26.6^{*}$ \\
 & $P_{\mathrm{thin}}$ & $56.5 \pm 8.8$ & {\boldmath$76.2 \pm 7.0$} & $+19.7^{*}$ & $57.5 \pm 21.3$ & {\boldmath$71.4 \pm 14.1$} & $+13.9^{*}$ \\
\end{tabular}
\end{ruledtabular}
\end{table*}

%% file: sections/05_conclusion.tex
\section{Conclusion}
\label{sec:conclusion}

We compared a plain ViT (\medsam{}), a hybrid (\litemedsam{}) and three CNNs on the thin vessels of three retinal datasets, DRIVE, HRF and FIVES, with VessMap as a check in a second imaging domain. Table~\ref{tab:medsam_vs_cnn} summarizes the raw segmentation performance, comparing fine-tuned \medsam{} with the best Dice and $\thinrecall$ achieved by a CNN on each dataset.

\input{figures/results/tab03_medsam_vs_cnn}

Fine-tuned from its pre-trained weights, \medsam{} had an $\thinrecall$ not significantly different from that of the best CNN on all four datasets, with differences from $-0.6$ to $+2.5$ percentage points. For \medsam{} in these conditions, the results do not support the expectation that a plain ViT misses thin vessels. Importantly, this holds for the default $16 \times 16$~px patch of \medsam{}, which is much wider than thin vessels.

The drawback is cost. The cost per image of \medsam{} was 2 to 18 times that of \convunext{} and more than 30 times that of \lwnet{}. A compact plain ViT trained from scratch did not keep the performance of the original model. Against the best CNN, its $\thinrecall$ was significantly lower on DRIVE and FIVES, and its $P_{\mathrm{thin}}$ and Dice were significantly lower on all four datasets. The hybrid \litemedsam{} had the highest mean $\thinrecall$ on every dataset, but a lower $P_{\mathrm{thin}}$ than the best CNNs. Where cost matters, compact CNNs remain good candidates on small datasets. \lwnet{}, with 0.07~M parameters, had a Dice and an $\thinrecall$ not significantly different from those of fine-tuned \medsam{} on DRIVE and VessMap.

Pre-training was essential for \medsam{}. Without it, \medsam{} fell significantly below the best CNN in $\thinrecall$ on three datasets. The Dice gain from pre-training was smallest on FIVES, the dataset with the most training images.

Our study has limitations that suggest directions for future work. First, the CNNs were trained from scratch, while the transformer-based models were fine-tuned. As mentioned in Section~\ref{subsec:method_training}, ImageNet weights exist for the ConvNeXt-Large encoder~\cite{liu2022convnet}, but smaller models, which are commonly used in practice, lack pre-trained encoders. Second, although the hyperparameters of each configuration were selected from 100 trials, the reported results come from a single training run, that of the best trial. Furthermore, the optimization protocol varied slightly among datasets and models, for instance in the validation interval and in the range of batch sizes (Appendix~\ref{app:training}). Some design choices, such as the optimizer and the learning-rate schedule, were also kept fixed.

Thin vessels are challenging to annotate, so metrics calculated on them can have high variability. The quality of biomedical annotations depends on factors such as the labeling instructions and the expertise of the annotators~\cite{radsch2023labelling}. On DRIVE, for instance, the $\thinrecall$ between the two annotators is $58.6\% \pm 11.9\%$, close to the value obtained for \medsam{} (56.9\%). However, the influence of the annotator is challenging to quantify, since most retinal datasets do not contain multiple annotations. This constitutes an interesting direction for future work.

\section*{CRediT authorship contribution statement}

Carlos Eduardo Fontaneli: Methodology, Investigation, Software, Data curation, Writing - original draft. Cesar H. Comin: Conceptualization, Supervision, Methodology, Writing - original draft, Writing - review and editing.

\section*{Funding}

This work was supported by the S\~ao Paulo Research Foundation (FAPESP) [grant number 25/04800-9].

\section*{Data availability}

The four datasets used in this study are publicly available from their original sources: DRIVE~\cite{Staal2004}, HRF~\cite{budai2013robust}, FIVES~\cite{jin2022fives} and VessMap~\cite{viana2025new}. The code of the experiments is available at \url{https://github.com/CarlosFontaneli/thin-vessel-sam-cnn}.

\section*{Declaration of generative AI and AI-assisted technologies in the manuscript preparation process}

During the preparation of this work the authors used Claude (Anthropic) in order to improve the language and flow of the text. After using this tool, the authors reviewed and edited the content as needed and take full responsibility for the content of the published article.

%% file: figures/results/tab03_medsam_vs_cnn.tex
\begin{table}[!htbp]
\caption{Fine-tuned \medsam{} against the best CNN of each metric and dataset. L: \lwnet{}, U: \convunext{}, N: \convnext{}. Bold indicates the best value. Asterisks indicate significantly different values.}
\label{tab:medsam_vs_cnn}
\footnotesize
\begin{ruledtabular}
\begin{tabular}{lr@{}lr@{}lr@{}lr@{}l}
 & \multicolumn{4}{c}{Dice} & \multicolumn{4}{c}{$\thinrecall$} \\
\cmidrule(lr){2-5} \cmidrule(lr){6-9}
Dataset & \multicolumn{2}{c}{\medsam{}} & \multicolumn{2}{c}{Best CNN} & \multicolumn{2}{c}{\medsam{}} & \multicolumn{2}{c}{Best CNN} \\
\midrule
DRIVE & {\boldmath$81.3 \pm {}$} & {\boldmath$1.3$} & $81.1 \pm {}$ & $2.4$\textsuperscript{L} & $56.9 \pm {}$ & $11.8$ & {\boldmath$57.5 \pm {}$} & {\boldmath$12.1$}\textsuperscript{L} \\
HRF & {\boldmath$81.7 \pm {}$} & {\boldmath$4.5^{*}$} & $79.7 \pm {}$ & $5.1$\textsuperscript{U} & {\boldmath$26.9 \pm {}$} & {\boldmath$17.2$} & $25.4 \pm {}$ & $13.5$\textsuperscript{N} \\
FIVES & {\boldmath$92.2 \pm {}$} & {\boldmath$8.5^{*}$} & $90.2 \pm {}$ & $8.5$\textsuperscript{N} & $36.2 \pm {}$ & $11.9$ & {\boldmath$36.5 \pm {}$} & {\boldmath$12.2$}\textsuperscript{N} \\
VessMap & $92.8 \pm {}$ & $2.1$ & {\boldmath$92.9 \pm {}$} & {\boldmath$2.1$}\textsuperscript{U} & {\boldmath$82.6 \pm {}$} & {\boldmath$14.1$} & $80.1 \pm {}$ & $12.2$\textsuperscript{L} \\
\end{tabular}
\end{ruledtabular}
\end{table}

%% file: sections/06_appendix.tex
\clearpage
\appendix

\counterwithin*{table}{section}
\counterwithin*{figure}{section}
\counterwithin*{equation}{section}
\renewcommand{\thetable}{\thesection.\arabic{table}}
\renewcommand{\thefigure}{\thesection.\arabic{figure}}
\renewcommand{\theequation}{\thesection.\arabic{equation}}

\section{Training details}
\label{app:training}

\subsection{Data preparation, sampling and augmentation}

Each training sample was flipped horizontally and vertically at random (probability 0.5 each) and rotated by a random angle between $-30^\circ$ and $30^\circ$. The sample was then cropped at a random position to the model's input size: 256~px for the CNNs, \litemedsam{} and the compact models, and 1024~px for \medsam{}. Images smaller than the input size, such as the DRIVE ($565 \times 584$~px) and VessMap ($256 \times 256$~px) images for \medsam{}, were upscaled to it with bilinear interpolation for the image and nearest-exact interpolation for the mask.

In 106 training masks and 5 test masks of FIVES, pixels along the image border were incorrectly labeled as vessel. We corrected this by setting a 2~px border of these masks to background.

\subsection{Loss function}
All models were trained with the soft Dice loss with squared terms in the denominator~\cite{milletari2016vnet},
\begin{equation}
    \mathcal{L}_{\text{Dice}}(Y, \hat{Y}) = 1 - \frac{2 \sum_{i,j} Y_{i,j} \hat{Y}_{i,j} + \epsilon}{\sum_{i,j} Y_{i,j}^2 + \sum_{i,j} \hat{Y}_{i,j}^2 + \epsilon},
    \label{eq:dice_loss}
\end{equation}
where $\hat{Y}$ is the sigmoid output and $\epsilon = 10^{-5}$. The loss was computed for each image and averaged over the batch. \lwnet{} produces a coarse and a refined segmentation, and its loss is the sum of Eq.~\eqref{eq:dice_loss} over the two outputs.

\subsection{Hyperparameter search}

Table~\ref{tab:search_space} gives the search space of the training hyperparameters and of the CNN architectures for each configuration and dataset. Every run used AdamW with a cosine learning-rate schedule~\cite{loshchilov2017sgdr} over at most 1000 epochs.

For each CNN, the search covered one architecture parameter. For \lwnet{} and \convunext{}, this parameter is the base width, the number of channels of the first level, which sets the size of the model. It was searched on DRIVE, FIVES and VessMap and fixed on HRF at 8 for \lwnet{} and 32 for \convunext{}. The selected \lwnet{} has a base width of 8 on every dataset (0.07~M parameters), and the selected \convunext{} has a base width of 64 on DRIVE and VessMap (13.8~M) and 32 on HRF and FIVES (3.5~M). For \convnext{}, the searched parameter is the stochastic-depth rate, which does not change its size (211.4~M).

For the compact models, the search also covered the architecture. For compact \medsam{}, it covered the patch size (1, 4, 8 or 16~px on DRIVE and 1, 2, 4, 8 or 16~px on the other datasets), the embedding dimension (128, 256 or 384), the depth (4, 6 or 8), the number of attention heads (4 or 8), the MLP ratio, that is, the ratio of the hidden width of the feed-forward block (2, 4 or 6), and the attention window (7 or 14). For compact \litemedsam{}, it covered the embedding dimensions of the four stages, (32, 64, 80, 160) or (64, 128, 160, 320), and their depths, (1, 1, 3, 1) or (2, 2, 6, 2). On DRIVE, it also covered the number of heads per stage, (1, 2, 2, 5) or (2, 4, 5, 10), the attention windows, (7, 7, 7, 7) or (7, 7, 14, 7), and the MLP ratio (2, 4 or 6). On the other datasets, these three parameters kept the values of the original \litemedsam{}. Table~\ref{tab:compact_models} contains the selected architectures. Table~\ref{tab:patch_size} gives the best validation Dice that the search reached with each patch size.

Validation ran the sliding-window inference of Section~\ref{subsec:method_metrics} over the full validation images. To reduce training time, every run on HRF, every compact model and the transformer-based models on FIVES were validated every 10 epochs, except the selected run of fine-tuned \litemedsam{} on FIVES (Table~\ref{tab:search_space}). The other runs were validated every epoch. In both cases, training stopped after 50 epochs without improvement. For \medsam{} on DRIVE and VessMap, validation used the padded protocol of Section~\ref{subsec:method_metrics}, while the test results of \medsam{} used the upscaled protocol, except in the comparison of Section~\ref{subsec:results_padding}.

\input{figures/results/tabA1_search_space}

\input{figures/results/tabA2_compact_models}

\input{figures/results/tabA3_patch_size}

\section{Pre-training results}
\label{app:results_detail}

Table~\ref{tab:pretraining_gain} gives the values plotted in Fig.~\ref{fig:pretraining}. It shows the Dice, $\thinrecall$ and $P_{\mathrm{thin}}$ of \medsam{} and \litemedsam{} fine-tuned and trained from scratch, the gain from pre-training and the values of the best CNN.

\input{figures/results/tab04_pretraining_gain}

\section{Statistical tests}
\label{app:statistics}
\input{figures/results/tabA4_pvalues}

Tables~\ref{tab:pvalues} and \ref{tab:pvalues_other} list the Holm-adjusted p-values of the pairwise comparisons of Section~\ref{sec:results_discussion}. The Holm correction was applied separately to each family of tests. In Table~\ref{tab:pvalues}, a family is formed by the 36 pairs of the nine configurations for one dataset and metric. In the top part of Table~\ref{tab:pvalues_other}, a family is formed by the 10 pairs of the five models for one dataset and metric. The 12 tests of the bottom part, which compare the two protocols of \medsam{}, form a single family. Every model was evaluated on the same test images of a dataset, so we compared models through their per-image paired differences. Zero differences were discarded, and an image with an undefined value was left out of the pairs that involve it. The p-values were computed with SciPy~\cite{virtanen2020scipy} (version 1.17.1), from the exact null distribution of the signed-rank statistic for up to 50 image pairs and from its normal approximation otherwise. The direction marks of the two tables follow the sign of the Hodges--Lehmann estimate~\cite{hodges1963estimates} of the paired difference, that is, the median of the averages of all pairs of per-image differences.

\clearpage

%% file: figures/results/tabA1_search_space.tex
\begin{table*}[tbp]
\caption{Hyperparameter search. Learning rate and weight decay: log-uniform in the ranges shown. CNN architecture parameter: \lwnet{}, width of the first of its three levels (doubled at each level); \convunext{}, width of the first stage; \convnext{}, stochastic-depth (drop-path) rate. Val.: validation interval in epochs. GPU~h: sum of the trial durations.}
\label{tab:search_space}
\footnotesize
\begin{ruledtabular}
\begin{tabular}{llcccccr}
Configuration & Dataset & Learning rate & Weight decay & Batch size & Architecture & Val. & GPU~h \\
\midrule
\multirow{4}{*}{\lwnet{}} & DRIVE & $10^{-6}$--$10^{-1}$ & $10^{-6}$--$10^{-1}$ & 2, 4, 8 & first width 8, 16, 32 & 1 & 13.0 \\
 & HRF & $10^{-6}$--$10^{-1}$ & $10^{-6}$--$10^{-1}$ & 2, 4, 8, 16 & -- & 10 & 34.0 \\
 & FIVES & $10^{-6}$--$10^{-1}$ & $10^{-6}$--$10^{-1}$ & 2, 4, 8, 16, 32, 64 & first width 8, 16, 32 & 1 & 108.6 \\
 & VessMap & $10^{-5}$--$10^{-2}$ & $10^{-5}$--$10^{-1}$ & 2, 4, 8, 16, 32 & first width 8, 16, 32 & 1 & 1.2 \\
\midrule
\multirow{4}{*}{\convunext{}} & DRIVE & $10^{-6}$--$10^{-2}$ & $10^{-6}$--$10^{-1}$ & 2, 4, 8 & base width 16, 32, 64 & 1 & 13.4 \\
 & HRF & $10^{-6}$--$10^{-2}$ & $10^{-6}$--$10^{-1}$ & 2, 4, 8, 16 & -- & 10 & 25.1 \\
 & FIVES & $10^{-6}$--$10^{-2}$ & $10^{-6}$--$10^{-1}$ & 2, 4, 8, 16, 32, 64 & base width 16, 32, 64 & 1 & 292.6 \\
 & VessMap & $10^{-5}$--$10^{-2}$ & $10^{-5}$--$10^{-1}$ & 2, 4, 8, 16, 32 & base width 16, 32, 64 & 1 & 2.5 \\
\midrule
\multirow{4}{*}{\convnext{}} & DRIVE & $10^{-6}$--$10^{-2}$ & $10^{-6}$--$10^{-1}$ & 2, 4, 8 & drop path 0--0.5 & 1 & 15.1 \\
 & HRF & $10^{-6}$--$10^{-2}$ & $10^{-6}$--$10^{-1}$ & 2, 4, 8, 16 & drop path 0--0.5 & 10 & 40.2 \\
 & FIVES & $10^{-6}$--$10^{-2}$ & $10^{-6}$--$10^{-1}$ & 2, 4, 8, 16, 32 & drop path 0--0.5 & 1 & 210.5 \\
 & VessMap & $10^{-5}$--$10^{-2}$ & $10^{-5}$--$10^{-1}$ & 2, 4, 8, 16, 32 & drop path 0--0.5 & 1 & 31.5 \\
\midrule
\multirow{4}{*}{\litemedsam{}, fine-tuned} & DRIVE & $10^{-6}$--$10^{-2}$ & $10^{-6}$--$10^{-1}$ & 2, 4, 8 & -- & 1 & 7.6 \\
 & HRF & $10^{-6}$--$10^{-2}$ & $10^{-6}$--$10^{-1}$ & 4, 8 & -- & 10 & 7.0 \\
 & FIVES & $10^{-6}$--$10^{-2}$ & $10^{-6}$--$10^{-1}$ & 8, 16 & -- & 1 & 117.3 \\
 & VessMap & $10^{-5}$--$10^{-2}$ & $10^{-5}$--$10^{-1}$ & 2, 4, 8 & -- & 1 & 20.9 \\
\midrule
\multirow{4}{*}{\litemedsam{}, from scratch} & DRIVE & $10^{-6}$--$10^{-2}$ & $10^{-6}$--$10^{-1}$ & 1, 2, 4 & -- & 1 & 17.4 \\
 & HRF & $10^{-6}$--$10^{-2}$ & $10^{-6}$--$10^{-1}$ & 4, 8 & -- & 10 & 13.6 \\
 & FIVES & $10^{-6}$--$10^{-2}$ & $10^{-6}$--$10^{-1}$ & 4, 8 & -- & 10 & 138.1 \\
 & VessMap & $10^{-5}$--$10^{-2}$ & $10^{-5}$--$10^{-1}$ & 1, 2, 4 & -- & 1 & 9.0 \\
\midrule
\multirow{4}{*}{\litemedsam{}, compact} & DRIVE & $10^{-6}$--$10^{-2}$ & $10^{-6}$--$10^{-1}$ & 8, 16, 32 & Table~\ref{tab:compact_models} & 10 & 1.7 \\
 & HRF & $10^{-6}$--$10^{-2}$ & $10^{-6}$--$10^{-1}$ & 8, 16, 32 & Table~\ref{tab:compact_models} & 10 & 11.6 \\
 & FIVES & $10^{-6}$--$10^{-2}$ & $10^{-6}$--$10^{-1}$ & 8, 16, 32 & Table~\ref{tab:compact_models} & 10 & 107.3 \\
 & VessMap & $10^{-5}$--$10^{-2}$ & $10^{-5}$--$10^{-1}$ & 8, 16, 32 & Table~\ref{tab:compact_models} & 10 & 17.2 \\
\midrule
\multirow{4}{*}{\medsam{}, fine-tuned} & DRIVE & $10^{-7}$--$10^{-3}$ & $10^{-6}$--$10^{-1}$ & 2 & -- & 1 & 9.9 \\
 & HRF & $10^{-7}$--$10^{-3}$ & $10^{-6}$--$10^{-1}$ & 2, 4 & -- & 10 & 15.0 \\
 & FIVES & $10^{-7}$--$10^{-3}$ & $10^{-6}$--$10^{-1}$ & 2, 4 & -- & 10 & 277.1 \\
 & VessMap & $10^{-5}$--$10^{-2}$ & $10^{-5}$--$10^{-1}$ & 2 & -- & 1 & 21.1 \\
\midrule
\multirow{4}{*}{\medsam{}, from scratch} & DRIVE & $10^{-7}$--$10^{-3}$ & $10^{-6}$--$10^{-1}$ & 1 & -- & 1 & 28.6 \\
 & HRF & $10^{-7}$--$10^{-3}$ & $10^{-6}$--$10^{-1}$ & 1, 2, 4 & -- & 10 & 20.8 \\
 & FIVES & $10^{-7}$--$10^{-3}$ & $10^{-6}$--$10^{-1}$ & 1, 2, 4 & -- & 10 & 345.2 \\
 & VessMap & $10^{-5}$--$10^{-2}$ & $10^{-5}$--$10^{-1}$ & 1 & -- & 1 & 26.3 \\
\midrule
\multirow{4}{*}{\medsam{}, compact} & DRIVE & $10^{-7}$--$10^{-3}$ & $10^{-6}$--$10^{-1}$ & 4, 8, 16 & Table~\ref{tab:compact_models} & 10 & 0.9 \\
 & HRF & $10^{-7}$--$10^{-3}$ & $10^{-6}$--$10^{-1}$ & 4, 8, 16 & Table~\ref{tab:compact_models} & 10 & 7.6 \\
 & FIVES & $10^{-7}$--$10^{-3}$ & $10^{-6}$--$10^{-1}$ & 4, 8, 16 & Table~\ref{tab:compact_models} & 10 & 66.5 \\
 & VessMap & $10^{-5}$--$10^{-2}$ & $10^{-5}$--$10^{-1}$ & 4, 8, 16 & Table~\ref{tab:compact_models} & 10 & 17.7 \\
\end{tabular}
\end{ruledtabular}
\end{table*}

%% file: figures/results/tabA2_compact_models.tex
\begin{table*}[tbp]
\caption{Selected compact models and, for comparison, the original \medsam{} and \litemedsam{}. Input and Patch in px. Tokens: patch grid of one window, calculated as image size divided by patch size. Emb.: embedding dimension. MLP: ratio of the hidden width of the MLP block. Window: size of the attention window in patches. Params: number of parameters in millions.}
\label{tab:compact_models}
\scriptsize
\begin{ruledtabular}
\begin{tabular}{llccccccccrccc}
Model & Version & Input & Patch & Tokens & Emb. & Depth & Heads & MLP & Window & Params & Dice & $\thinrecall$ & $P_{\mathrm{thin}}$ \\
\midrule
\multirow{5}{*}{\medsam{}} & original & 1024 & 16 & $64\times64$ & 768 & 12 & 12 & 4 & 14 & 93.7 & \multicolumn{3}{c}{Table~\ref{tab:pretraining_gain}} \\
 & compact, DRIVE & 256 & 4 & $64\times64$ & 256 & 4 & 8 & 2 & 14 & 7.9 & $77.7{\pm}2.5$ & $43.4{\pm}13.5$ & $69.3{\pm}8.3$ \\
 & compact, HRF & 256 & 4 & $64\times64$ & 128 & 6 & 4 & 2 & 14 & 6.0 & $78.3{\pm}5.9$ & $23.7{\pm}15.3$ & $43.3{\pm}11.6$ \\
 & compact, FIVES & 256 & 8 & $32\times32$ & 256 & 8 & 8 & 6 & 7 & 13.5 & $87.8{\pm}9.0$ & $32.0{\pm}11.3$ & $41.9{\pm}10.7$ \\
 & compact, VessMap & 256 & 4 & $64\times64$ & 384 & 6 & 4 & 6 & 7 & 20.6 & $91.7{\pm}2.8$ & $77.5{\pm}18.2$ & $74.5{\pm}14.5$ \\
\midrule
\multirow{5}{*}{\litemedsam{}} & original & 256 & -- & -- & 64,128,160,320 & 2,2,6,2 & 2,4,5,10 & 4 & 7,7,14,7 & 9.8 & \multicolumn{3}{c}{Table~\ref{tab:pretraining_gain}} \\
 & compact, DRIVE & 256 & -- & -- & 64,128,160,320 & 2,2,6,2 & 2,4,5,10 & 6 & 7,7,7,7 & 11.4 & $81.4{\pm}2.0$ & $62.4{\pm}12.1$ & $72.9{\pm}5.7$ \\
 & compact, HRF & 256 & -- & -- & 32,64,80,160 & 1,1,3,1 & 2,4,5,10 & 4 & 7,7,14,7 & 5.4 & $79.2{\pm}5.7$ & $30.5{\pm}13.9$ & $41.2{\pm}11.9$ \\
 & compact, FIVES & 256 & -- & -- & 64,128,160,320 & 1,1,3,1 & 2,4,5,10 & 4 & 7,7,14,7 & 7.4 & $86.7{\pm}10.8$ & $32.1{\pm}10.7$ & $41.2{\pm}12.1$ \\
 & compact, VessMap & 256 & -- & -- & 64,128,160,320 & 1,1,3,1 & 2,4,5,10 & 4 & 7,7,14,7 & 7.4 & $92.3{\pm}1.9$ & $81.8{\pm}11.5$ & $75.7{\pm}15.6$ \\
\end{tabular}
\end{ruledtabular}
\end{table*}

%% file: figures/results/tabA3_patch_size.tex
\begin{table}[tbp]
\caption{Patch sizes in the search of compact \medsam{}. Number of trials that reached a validation epoch (Tr.) and their best validation Dice (\%). Bold indicates the best performance for each dataset, which set the patch sizes used in the experiments. OOM: every trial ran out of GPU memory.}
\label{tab:patch_size}
\begin{ruledtabular}
\begin{tabular}{crrrrrrrr}
 & \multicolumn{2}{c}{DRIVE} & \multicolumn{2}{c}{HRF} & \multicolumn{2}{c}{FIVES} & \multicolumn{2}{c}{VessMap} \\
\cmidrule(lr){2-3} \cmidrule(lr){4-5} \cmidrule(lr){6-7} \cmidrule(lr){8-9}
Patch (px) & Tr. & Dice & Tr. & Dice & Tr. & Dice & Tr. & Dice \\
\midrule
1 & 0 & OOM & 0 & OOM & 0 & OOM & 0 & OOM \\
2 & 0 & OOM & 0 & OOM & 0 & OOM & 0 & OOM \\
4 & 18 & \textbf{76.8} & 44 & \textbf{78.9} & 4 & 85.3 & 37 & \textbf{92.1} \\
8 & 16 & 70.2 & 13 & 75.0 & 69 & \textbf{89.6} & 32 & 90.5 \\
16 & 56 & 15.7 & 16 & 14.0 & 8 & 82.7 & 6 & 87.0 \\
\end{tabular}
\end{ruledtabular}
\end{table}

%% file: figures/results/tab04_pretraining_gain.tex
\begin{table*}[tbp]
\caption{Effect of pre-training. \litemedsam{} and \medsam{} fine-tuned (FT) and trained from scratch, and the best CNN. L: \lwnet{}, U: \convunext{}, N: \convnext{}. $\Delta$ shows the FT value minus Scratch, with an asterisk when the difference is significant. Bold indicates the best value, including the CNN. Underlined values are not significantly different from the best of the same row.}
\label{tab:pretraining_gain}
\begin{ruledtabular}
\begin{tabular}{llccccccc}
 & & \multicolumn{3}{c}{\litemedsam{}} & \multicolumn{3}{c}{\medsam{}} & \\
\cmidrule(lr){3-5} \cmidrule(lr){6-8}
Dataset & Metric & FT & Scratch & $\Delta$ & FT & Scratch & $\Delta$ & CNN \\
\midrule
\multirow{3}{*}{\shortstack{DRIVE\\($n=20$)}} & Dice & {\boldmath$82.3 \pm 1.9$} & $81.4 \pm 2.4$ & $+0.9^{*}$ & $81.3 \pm 1.3$ & $78.7 \pm 1.6$ & $+2.6^{*}$ & $81.1 \pm 2.4$\textsuperscript{L} \\
 & $\thinrecall$ & {\boldmath$64.3 \pm 11.6$} & $59.9 \pm 13.4$ & $+4.4^{*}$ & $56.9 \pm 11.8$ & $39.3 \pm 11.2$ & $+17.6^{*}$ & $57.5 \pm 12.1$\textsuperscript{L} \\
 & $P_{\mathrm{thin}}$ & $74.5 \pm 5.2$ & $76.2 \pm 5.8$ & $-1.7$ & $73.0 \pm 6.1$ & \underline{$76.2 \pm 7.0$} & $-3.2^{*}$ & {\boldmath$80.8 \pm 8.6$}\textsuperscript{U} \\
\midrule
\multirow{3}{*}{\shortstack{HRF\\($n=14$)}} & Dice & \underline{$80.6 \pm 5.9$} & $79.5 \pm 5.1$ & $+1.1$ & {\boldmath$81.7 \pm 4.5$} & $79.7 \pm 5.4$ & $+2.0^{*}$ & $79.7 \pm 5.1$\textsuperscript{U} \\
 & $\thinrecall$ & {\boldmath$40.1 \pm 15.7$} & $30.2 \pm 13.6$ & $+9.9^{*}$ & $26.9 \pm 17.2$ & $28.0 \pm 16.1$ & $-1.1$ & $25.4 \pm 13.5$\textsuperscript{N} \\
 & $P_{\mathrm{thin}}$ & $35.7 \pm 12.1$ & $46.9 \pm 13.0$ & $-11.2^{*}$ & \underline{$55.6 \pm 14.6$} & $41.2 \pm 12.7$ & $+14.4^{*}$ & {\boldmath$56.2 \pm 15.7$}\textsuperscript{U} \\
\midrule
\multirow{3}{*}{\shortstack{FIVES\\($n=200$)}} & Dice & $88.1 \pm 9.8$ & $86.6 \pm 11.1$ & $+1.5^{*}$ & {\boldmath$92.2 \pm 8.5$} & $91.6 \pm 8.9$ & $+0.6^{*}$ & $90.2 \pm 8.5$\textsuperscript{N} \\
 & $\thinrecall$ & {\boldmath$37.3 \pm 12.8$} & $27.1 \pm 11.3$ & $+10.2^{*}$ & \underline{$36.2 \pm 11.9$} & $34.3 \pm 11.8$ & $+1.9^{*}$ & \underline{$36.5 \pm 12.2$}\textsuperscript{N} \\
 & $P_{\mathrm{thin}}$ & $38.3 \pm 13.7$ & $47.6 \pm 14.0$ & $-9.3^{*}$ & {\boldmath$51.7 \pm 15.7$} & \underline{$50.2 \pm 14.4$} & $+1.5$ & \underline{$49.7 \pm 15.8$}\textsuperscript{L} \\
\midrule
\multirow{3}{*}{\shortstack{VessMap\\($n=30$)}} & Dice & {\boldmath$93.0 \pm 2.0$} & $91.0 \pm 3.6$ & $+2.0^{*}$ & \underline{$92.8 \pm 2.1$} & $89.7 \pm 3.7$ & $+3.1^{*}$ & \underline{$92.9 \pm 2.1$}\textsuperscript{U} \\
 & $\thinrecall$ & {\boldmath$83.8 \pm 12.1$} & $72.3 \pm 18.7$ & $+11.5^{*}$ & \underline{$82.6 \pm 14.1$} & $70.2 \pm 20.9$ & $+12.4^{*}$ & \underline{$80.1 \pm 12.2$}\textsuperscript{L} \\
 & $P_{\mathrm{thin}}$ & \underline{$81.9 \pm 14.5$} & {\boldmath$84.6 \pm 15.4$} & $-2.7$ & $79.6 \pm 14.8$ & $71.4 \pm 14.1$ & $+8.2^{*}$ & \underline{$83.8 \pm 12.9$}\textsuperscript{U} \\
\end{tabular}
\end{ruledtabular}
\end{table*}

%% file: figures/results/tabA4_pvalues.tex
\begin{table*}[tbp]
\caption{Holm-adjusted p-values of the pairwise tests. A $+$ superscript indicates that the first model of the pair is significantly better ($p < 0.05$). A $-$ is used when the second configuration is better. LW: \lwnet{}; CU: \convunext{}; CN: \convnext{}; LM, LM-S, LM-C: \litemedsam{} fine-tuned, from scratch and compact; M, M-S, M-C: \medsam{} fine-tuned, from scratch and compact.}
\label{tab:pvalues}
\scriptsize
\begin{ruledtabular}
\begin{tabular}{lcccccccccccc}
 & \multicolumn{3}{c}{DRIVE} & \multicolumn{3}{c}{HRF} & \multicolumn{3}{c}{FIVES} & \multicolumn{3}{c}{VessMap} \\
\cmidrule(lr){2-4} \cmidrule(lr){5-7} \cmidrule(lr){8-10} \cmidrule(lr){11-13}
Pair & Dice & $\thinrecall$ & $P_{\mathrm{thin}}$ & Dice & $\thinrecall$ & $P_{\mathrm{thin}}$ & Dice & $\thinrecall$ & $P_{\mathrm{thin}}$ & Dice & $\thinrecall$ & $P_{\mathrm{thin}}$ \\
\midrule
LW vs CU & 0.58 & 0.054 & 0.19 & 0.004$^{-}$ & 1.00 & 0.54 & $<$0.001$^{-}$ & $<$0.001$^{-}$ & $<$0.001$^{+}$ & 1.00 & 1.00 & 1.00 \\
LW vs CN & $<$0.001$^{+}$ & $<$0.001$^{+}$ & 0.010$^{+}$ & 1.00 & 1.00 & 0.59 & $<$0.001$^{-}$ & $<$0.001$^{-}$ & 0.059 & $<$0.001$^{+}$ & $<$0.001$^{+}$ & 0.75 \\
LW vs LM & 0.005$^{-}$ & $<$0.001$^{-}$ & $<$0.001$^{+}$ & 0.005$^{-}$ & 0.004$^{-}$ & 0.25 & $<$0.001$^{-}$ & $<$0.001$^{-}$ & $<$0.001$^{+}$ & 1.00 & 0.076 & 1.00 \\
LW vs LM-S & 0.44 & 0.008$^{-}$ & 0.025$^{+}$ & 0.004$^{-}$ & 0.62 & 1.00 & $<$0.001$^{-}$ & 0.33 & 1.00 & $<$0.001$^{+}$ & 0.002$^{+}$ & 0.15 \\
LW vs LM-C & 0.77 & $<$0.001$^{-}$ & $<$0.001$^{+}$ & 0.005$^{-}$ & 0.47 & 1.00 & $<$0.001$^{-}$ & $<$0.001$^{-}$ & $<$0.001$^{+}$ & 0.011$^{+}$ & 1.00 & 0.073 \\
LW vs M & 1.00 & 0.95 & $<$0.001$^{+}$ & 0.004$^{-}$ & 1.00 & 1.00 & $<$0.001$^{-}$ & $<$0.001$^{-}$ & 0.12 & 1.00 & 0.49 & 1.00 \\
LW vs M-S & $<$0.001$^{+}$ & $<$0.001$^{+}$ & 0.35 & 0.004$^{-}$ & 1.00 & 1.00 & $<$0.001$^{-}$ & $<$0.001$^{-}$ & 1.00 & $<$0.001$^{+}$ & $<$0.001$^{+}$ & 0.017$^{+}$ \\
LW vs M-C & $<$0.001$^{+}$ & $<$0.001$^{+}$ & 0.009$^{+}$ & 0.63 & 1.00 & 1.00 & $<$0.001$^{-}$ & $<$0.001$^{-}$ & $<$0.001$^{+}$ & $<$0.001$^{+}$ & 1.00 & 0.015$^{+}$ \\
CU vs CN & 0.012$^{+}$ & 0.007$^{+}$ & $<$0.001$^{+}$ & 0.005$^{+}$ & 1.00 & 0.004$^{+}$ & $<$0.001$^{-}$ & $<$0.001$^{-}$ & $<$0.001$^{-}$ & $<$0.001$^{+}$ & $<$0.001$^{+}$ & 0.002$^{+}$ \\
CU vs LM & $<$0.001$^{-}$ & $<$0.001$^{-}$ & 0.016$^{+}$ & 0.49 & 0.004$^{-}$ & 0.004$^{+}$ & $<$0.001$^{+}$ & $<$0.001$^{-}$ & 0.041$^{-}$ & 1.00 & $<$0.001$^{-}$ & 1.00 \\
CU vs LM-S & 0.026$^{-}$ & $<$0.001$^{-}$ & 0.012$^{+}$ & 1.00 & 0.010$^{-}$ & 0.015$^{+}$ & $<$0.001$^{+}$ & $<$0.001$^{+}$ & $<$0.001$^{-}$ & $<$0.001$^{+}$ & 0.002$^{+}$ & 1.00 \\
CU vs LM-C & 0.15 & $<$0.001$^{-}$ & 0.004$^{+}$ & 0.63 & 0.031$^{-}$ & 0.004$^{+}$ & $<$0.001$^{+}$ & 0.017$^{+}$ & $<$0.001$^{-}$ & $<$0.001$^{+}$ & 0.44 & $<$0.001$^{+}$ \\
CU vs M & 1.00 & 0.098 & 0.013$^{+}$ & 0.004$^{-}$ & 0.28 & 1.00 & $<$0.001$^{-}$ & 0.002$^{-}$ & $<$0.001$^{-}$ & 1.00 & 0.028$^{-}$ & 0.048$^{+}$ \\
CU vs M-S & 0.017$^{+}$ & 0.007$^{+}$ & 0.084 & 1.00 & 0.022$^{-}$ & 0.004$^{+}$ & $<$0.001$^{-}$ & 1.00 & $<$0.001$^{-}$ & $<$0.001$^{+}$ & $<$0.001$^{+}$ & 0.001$^{+}$ \\
CU vs M-C & 0.012$^{+}$ & 0.006$^{+}$ & $<$0.001$^{+}$ & 0.014$^{+}$ & 1.00 & 0.004$^{+}$ & $<$0.001$^{+}$ & 0.001$^{+}$ & $<$0.001$^{-}$ & $<$0.001$^{+}$ & 1.00 & $<$0.001$^{+}$ \\
CN vs LM & $<$0.001$^{-}$ & $<$0.001$^{-}$ & 0.049$^{-}$ & 0.004$^{-}$ & 0.004$^{-}$ & 1.00 & $<$0.001$^{+}$ & 0.27 & $<$0.001$^{+}$ & $<$0.001$^{-}$ & $<$0.001$^{-}$ & 0.20 \\
CN vs LM-S & $<$0.001$^{-}$ & $<$0.001$^{-}$ & $<$0.001$^{-}$ & 0.004$^{-}$ & 0.004$^{-}$ & 0.007$^{-}$ & $<$0.001$^{+}$ & $<$0.001$^{+}$ & 0.041$^{-}$ & 0.40 & 0.88 & 0.007$^{-}$ \\
CN vs LM-C & $<$0.001$^{-}$ & $<$0.001$^{-}$ & 0.19 & 0.004$^{-}$ & 0.031$^{-}$ & 0.009$^{-}$ & $<$0.001$^{+}$ & $<$0.001$^{+}$ & $<$0.001$^{+}$ & $<$0.001$^{-}$ & $<$0.001$^{-}$ & 1.00 \\
CN vs M & $<$0.001$^{-}$ & $<$0.001$^{-}$ & 0.13 & 0.004$^{-}$ & 1.00 & 0.004$^{-}$ & $<$0.001$^{-}$ & 1.00 & $<$0.001$^{-}$ & $<$0.001$^{-}$ & $<$0.001$^{-}$ & 1.00 \\
CN vs M-S & $<$0.001$^{-}$ & 0.022$^{+}$ & 0.025$^{-}$ & 0.004$^{-}$ & 0.67 & 0.21 & $<$0.001$^{-}$ & 0.002$^{+}$ & $<$0.001$^{-}$ & $<$0.001$^{+}$ & 1.00 & 0.52 \\
CN vs M-C & 0.81 & 0.95 & 1.00 & 0.004$^{-}$ & 1.00 & 0.015$^{-}$ & $<$0.001$^{+}$ & $<$0.001$^{+}$ & $<$0.001$^{+}$ & 0.012$^{-}$ & 0.002$^{-}$ & 1.00 \\
LM vs LM-S & 0.003$^{+}$ & $<$0.001$^{+}$ & 0.084 & 0.24 & 0.004$^{+}$ & 0.007$^{-}$ & $<$0.001$^{+}$ & $<$0.001$^{+}$ & $<$0.001$^{-}$ & $<$0.001$^{+}$ & $<$0.001$^{+}$ & 0.77 \\
LM vs LM-C & $<$0.001$^{+}$ & 0.004$^{+}$ & 0.084 & 0.035$^{+}$ & 0.004$^{+}$ & 0.024$^{-}$ & $<$0.001$^{+}$ & $<$0.001$^{+}$ & $<$0.001$^{-}$ & $<$0.001$^{+}$ & 0.35 & 0.004$^{+}$ \\
LM vs M & 0.011$^{+}$ & $<$0.001$^{+}$ & 0.35 & 0.49 & 0.004$^{+}$ & 0.007$^{-}$ & $<$0.001$^{-}$ & 0.27 & $<$0.001$^{-}$ & 0.94 & 1.00 & 1.00 \\
LM vs M-S & $<$0.001$^{+}$ & $<$0.001$^{+}$ & 0.35 & 0.46 & 0.004$^{+}$ & 0.25 & $<$0.001$^{-}$ & $<$0.001$^{+}$ & $<$0.001$^{-}$ & $<$0.001$^{+}$ & $<$0.001$^{+}$ & 0.018$^{+}$ \\
LM vs M-C & $<$0.001$^{+}$ & $<$0.001$^{+}$ & 0.072 & 0.006$^{+}$ & 0.004$^{+}$ & 0.12 & $<$0.001$^{+}$ & $<$0.001$^{+}$ & $<$0.001$^{-}$ & $<$0.001$^{+}$ & 0.077 & 0.081 \\
LM-S vs LM-C & 1.00 & 0.004$^{-}$ & $<$0.001$^{+}$ & 1.00 & 1.00 & 0.12 & 0.69 & $<$0.001$^{-}$ & $<$0.001$^{+}$ & 0.047$^{-}$ & $<$0.001$^{-}$ & 0.004$^{+}$ \\
LM-S vs M & 0.56 & 0.028$^{+}$ & 0.072 & 0.004$^{-}$ & 1.00 & 0.024$^{-}$ & $<$0.001$^{-}$ & $<$0.001$^{-}$ & 0.017$^{-}$ & $<$0.001$^{-}$ & $<$0.001$^{-}$ & 0.016$^{+}$ \\
LM-S vs M-S & 0.005$^{+}$ & $<$0.001$^{+}$ & 1.00 & 1.00 & 1.00 & 0.015$^{+}$ & $<$0.001$^{-}$ & $<$0.001$^{-}$ & 0.36 & $<$0.001$^{+}$ & 0.98 & $<$0.001$^{+}$ \\
LM-S vs M-C & $<$0.001$^{+}$ & $<$0.001$^{+}$ & 0.003$^{+}$ & 0.035$^{+}$ & 0.15 & 0.59 & $<$0.001$^{-}$ & $<$0.001$^{-}$ & $<$0.001$^{+}$ & 1.00 & 0.52 & $<$0.001$^{+}$ \\
LM-C vs M & 0.56 & 0.004$^{+}$ & 1.00 & 0.004$^{-}$ & 1.00 & 0.015$^{-}$ & $<$0.001$^{-}$ & $<$0.001$^{-}$ & $<$0.001$^{-}$ & 0.003$^{-}$ & 1.00 & 1.00 \\
LM-C vs M-S & $<$0.001$^{+}$ & $<$0.001$^{+}$ & 0.11 & 0.49 & 1.00 & 1.00 & $<$0.001$^{-}$ & 0.029$^{-}$ & $<$0.001$^{-}$ & $<$0.001$^{+}$ & $<$0.001$^{+}$ & 1.00 \\
LM-C vs M-C & $<$0.001$^{+}$ & $<$0.001$^{+}$ & 0.34 & 0.040$^{+}$ & 0.28 & 1.00 & $<$0.001$^{-}$ & 1.00 & 1.00 & 0.53 & 1.00 & 1.00 \\
M vs M-S & $<$0.001$^{+}$ & $<$0.001$^{+}$ & 0.010$^{-}$ & 0.004$^{+}$ & 1.00 & 0.004$^{+}$ & $<$0.001$^{+}$ & $<$0.001$^{+}$ & 0.36 & $<$0.001$^{+}$ & $<$0.001$^{+}$ & 0.046$^{+}$ \\
M vs M-C & $<$0.001$^{+}$ & $<$0.001$^{+}$ & 0.14 & 0.004$^{+}$ & 0.67 & 0.007$^{+}$ & $<$0.001$^{+}$ & $<$0.001$^{+}$ & $<$0.001$^{+}$ & $<$0.001$^{+}$ & 0.002$^{+}$ & 0.38 \\
M-S vs M-C & 0.006$^{+}$ & 0.022$^{-}$ & 0.027$^{+}$ & 0.005$^{+}$ & 0.053 & 1.00 & $<$0.001$^{+}$ & 0.003$^{+}$ & $<$0.001$^{+}$ & $<$0.001$^{-}$ & $<$0.001$^{-}$ & 1.00 \\
\end{tabular}
\end{ruledtabular}
\end{table*}

\begin{table*}[tbp]
\caption{Holm-adjusted p-values for Table~\ref{tab:global_baseline} (top) and Table~\ref{tab:padding} (bottom). LW: \lwnet{}; CU: \convunext{}; CN: \convnext{}; LM: \litemedsam{} fine-tuned; M: \medsam{} fine-tuned. See the caption of Table~\ref{tab:pvalues} for the meaning of the $+$ and $-$ superscripts.}
\label{tab:pvalues_other}
\footnotesize
\begin{ruledtabular}
\begin{tabular}{lcccccccc}
 & \multicolumn{2}{c}{DRIVE} & \multicolumn{2}{c}{HRF} & \multicolumn{2}{c}{FIVES} & \multicolumn{2}{c}{VessMap} \\
\cmidrule(lr){2-3} \cmidrule(lr){4-5} \cmidrule(lr){6-7} \cmidrule(lr){8-9}
Pair & $R_{\mathrm{cl}}$ & $P_{\mathrm{cl}}$ & $R_{\mathrm{cl}}$ & $P_{\mathrm{cl}}$ & $R_{\mathrm{cl}}$ & $P_{\mathrm{cl}}$ & $R_{\mathrm{cl}}$ & $P_{\mathrm{cl}}$ \\
\midrule
LW vs CU & 0.048$^{+}$ & 0.36 & 1.00 & 0.23 & $<$0.001$^{-}$ & $<$0.001$^{-}$ & 0.025$^{+}$ & 0.29 \\
LW vs CN & $<$0.001$^{+}$ & 1.00 & 1.00 & 0.026$^{+}$ & $<$0.001$^{-}$ & $<$0.001$^{-}$ & $<$0.001$^{+}$ & 1.00 \\
LW vs LM & $<$0.001$^{-}$ & $<$0.001$^{+}$ & 0.001$^{-}$ & 1.00 & $<$0.001$^{-}$ & $<$0.001$^{-}$ & 0.002$^{-}$ & 0.084 \\
LW vs M & 0.70 & $<$0.001$^{+}$ & 0.043$^{-}$ & 0.27 & $<$0.001$^{-}$ & $<$0.001$^{-}$ & 0.037$^{-}$ & 1.00 \\
CU vs CN & 0.003$^{+}$ & 0.43 & 1.00 & 0.001$^{+}$ & $<$0.001$^{-}$ & 0.14 & 0.028$^{+}$ & 0.55 \\
CU vs LM & $<$0.001$^{-}$ & 0.005$^{+}$ & 0.001$^{-}$ & 0.001$^{+}$ & $<$0.001$^{-}$ & $<$0.001$^{+}$ & $<$0.001$^{-}$ & 1.00 \\
CU vs M & 0.008$^{-}$ & 0.005$^{+}$ & 0.001$^{-}$ & 1.00 & $<$0.001$^{-}$ & $<$0.001$^{-}$ & $<$0.001$^{-}$ & 0.043$^{+}$ \\
CN vs LM & $<$0.001$^{-}$ & 0.022$^{+}$ & 0.001$^{-}$ & 0.004$^{-}$ & $<$0.001$^{+}$ & $<$0.001$^{+}$ & $<$0.001$^{-}$ & 0.55 \\
CN vs M & $<$0.001$^{-}$ & 0.024$^{+}$ & 0.001$^{-}$ & 0.001$^{-}$ & $<$0.001$^{-}$ & $<$0.001$^{-}$ & $<$0.001$^{-}$ & 1.00 \\
LM vs M & 0.002$^{+}$ & 1.00 & 0.001$^{+}$ & 0.001$^{-}$ & $<$0.001$^{-}$ & $<$0.001$^{-}$ & 0.69 & 0.29 \\
\end{tabular}
\end{ruledtabular}
\vspace{0.8em}
\begin{ruledtabular}
\begin{tabular}{lcccccc}
 & \multicolumn{3}{c}{DRIVE} & \multicolumn{3}{c}{VessMap} \\
\cmidrule(lr){2-4} \cmidrule(lr){5-7}
\medsam{} & Dice & $\thinrecall$ & $P_{\mathrm{thin}}$ & Dice & $\thinrecall$ & $P_{\mathrm{thin}}$ \\
\midrule
Fine-tuned & $<$0.001$^{+}$ & $<$0.001$^{+}$ & $<$0.001$^{+}$ & $<$0.001$^{+}$ & $<$0.001$^{+}$ & 0.35 \\
From scratch & $<$0.001$^{+}$ & $<$0.001$^{+}$ & $<$0.001$^{+}$ & $<$0.001$^{+}$ & $<$0.001$^{+}$ & 0.003$^{+}$ \\
\end{tabular}
\end{ruledtabular}
\end{table*}